\documentclass[11pt]{article}

\usepackage[final]{acl}

\usepackage{times}
\usepackage{latexsym}
\usepackage{booktabs}
\usepackage{multirow}
\usepackage{graphicx}
\usepackage[T1]{fontenc}

\usepackage[utf8]{inputenc}

\usepackage{microtype}

\usepackage{inconsolata}

\usepackage{graphicx}

\usepackage{hyperref}
\usepackage{url}

\usepackage{booktabs}   
\usepackage{caption} 
\usepackage{amsfonts}       
\usepackage{amsmath,amssymb}
\usepackage{nicefrac}       
\usepackage{microtype}      
\usepackage{xcolor}         
\usepackage{amsmath,calc}
\usepackage{enumitem}
\usepackage{graphicx} 
\usepackage{comment}
\usepackage[most]{tcolorbox}
\usepackage{makecell}
\usepackage{lipsum}
\usepackage{multirow}
\usepackage{subcaption}
\usepackage{colortbl}
\usepackage{arydshln}
\usepackage{wrapfig}
\usepackage{floatflt}

\usepackage{xspace}

\title{Revisiting Anthropomorphic Reflection Markers \\in Large Language Model Reasoning}

\author{
 \textbf{Yahan Yu},
 \textbf{Noa Nakanishi},
 \textbf{Fei Cheng},
\\
 Kyoto University, Japan 
\\
\texttt{yahan@nlp.ist.i.kyoto-u.ac.jp}, \texttt{nakanishi.noa.25z@st.kyoto-u.ac.jp}, \\ \texttt{feicheng@i.kyoto-u.ac.jp}}

\begin{document}
\maketitle
\begin{abstract}
Large Language Models (LLMs) often produce explicit reflective traces during complex reasoning, accompanied by anthropomorphic markers such as \textit{wait}, \textit{hmm}, and \textit{alternatively}. Although these markers are commonly used as visible indicators of reflection, their mechanisms remain unclear, which leaves the risk of overthinking associated with redundant and repetitive reflection markers. In this work, we revisit anthropomorphic reflection markers, examining their necessity for reasoning and role in the reflection. We suppress these markers through prompt-level and token-level interventions, and analyze their effects on task performance across four benchmarks and two model scales. Our results show that anthropomorphic markers are not uniformly necessary for reasoning performance: suppressing them can preserve or improve performance in several settings, especially under larger sampling budgets. Meanwhile, marker suppression does not necessarily remove reflection behavior, as models can still perform marker-free verification. These suggest that anthropomorphic markers tend to be surface cues rather than reliable proxies for reflection itself, and motivate future research on reasoning mechanisms beyond explicit marker patterns.
\end{abstract}

\section{Introduction}
\label{sec:introduction}

Large Language Models (LLMs) have shown strong performance on complex reasoning tasks when they are encouraged to produce intermediate reasoning steps~\citep{wei2022chain,kojima2022large,wang2023selfconsistency,zhou2023least,yao2023tree}. More recently, DeepSeek-R1~\citep{guo2025deepseek} has shown that reasoning models can generate long reasoning traces that include explicit self-reflection associated with anthropomorphic markers \citep{weng2023selfverification} such as \textit{wait}, \textit{hmm}, \textit{oh}, and \textit{alternatively}, thereby contributing to reasoning performance.

\begin{figure}[t]
    \centering
    \includegraphics[width=0.99\linewidth]{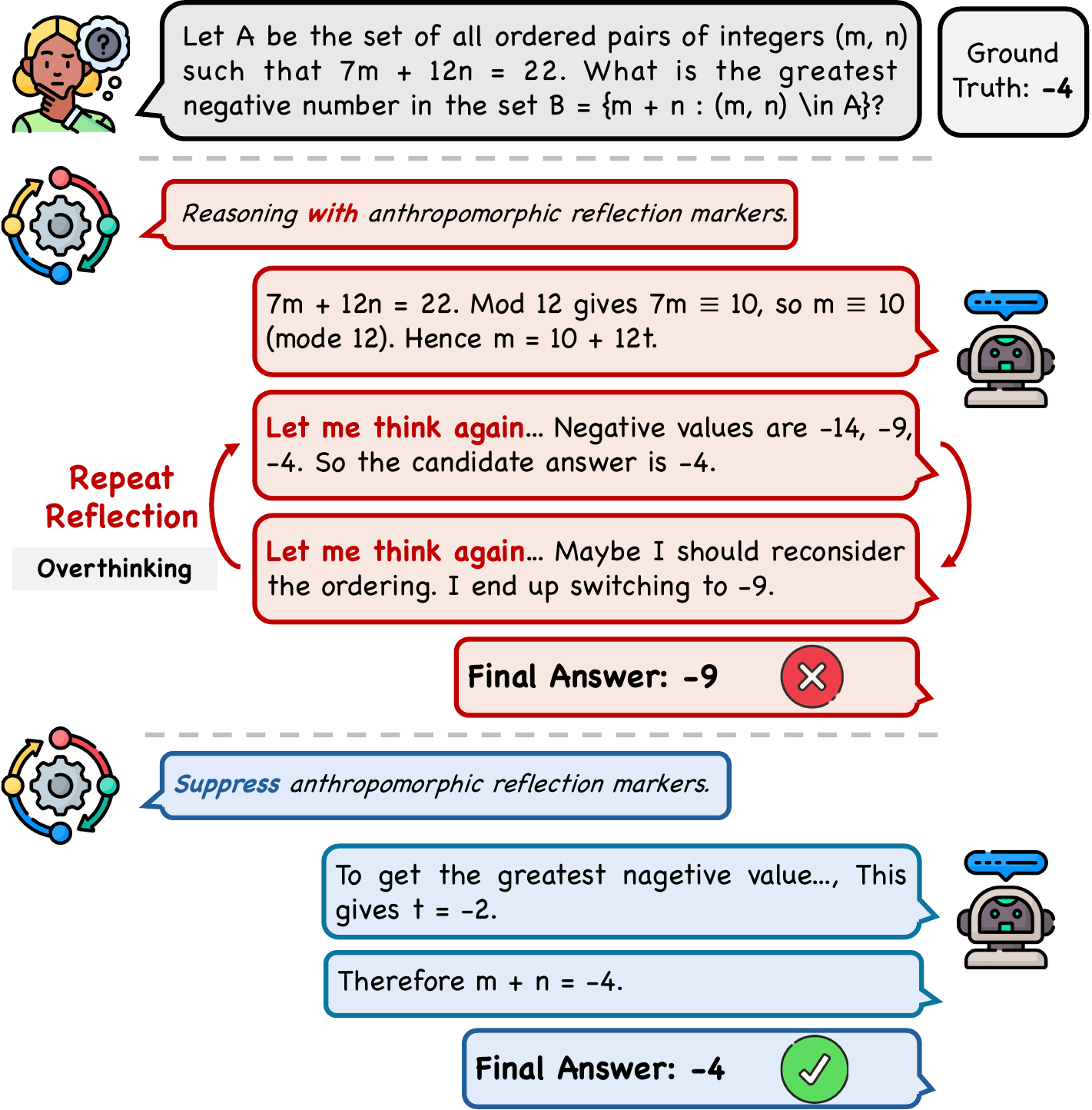}
    \caption{The overview of overthinking and the suppression of anthropomorphic reflection markers. Repeated markers may lead the model away from a correct intermediate answer, while marker suppression can mitigate these unnecessary verifications.}
    \label{fig:top}
\end{figure}

However, the mechanism behind anthropomorphic reflection markers remains underexplored. Prior work \citep{madaan2023selfrefine,shinn2023reflexion,zelikman2022star} has studied reflection mostly from a performance-oriented perspective, asking whether reflection improves final accuracy, but has paid less attention to what these markers actually do during generation. This gap is important because anthropomorphic markers may not only indicate reflective reasoning, but also shape the continuation of reasoning itself \citep{turpin2023language,lanham2023measuring}. When a model repeatedly produces markers such as \textit{wait}, \textit{let me check}, or \textit{I made a mistake}, these expressions may encourage additional verification even when the current reasoning path is already sufficient. As a result, reflection markers may provide a possible channel for overthinking \citep{huang2024cannot,li2024hindsight} and harm the performance \citep{chen2024donot,wang2025wait,fu2025reasoning}.



Therefore, we revisit anthropomorphic reflection markers from this perspective. We examine their necessity for reasoning and role in the reflection. This formulation leads to four concrete \textit{Research Questions (RQs)}: 
\begin{tcolorbox}
\textit{\textbf{RQ1}: Are the anthropomorphic markers necessary for problem-solving ability?}

\textit{\textbf{RQ2}: Are the observed effects specific to anthropomorphic reflection markers?}

\textit{\textbf{RQ3}: How closely are anthropomorphic markers associated with explicit reflection behavior?}

\textit{\textbf{RQ4}: For the mechanism of anthropomorphic markers, what can be revealed about?}
\end{tcolorbox}

Specifically, we investigate this problem by suppressing anthropomorphic reflection markers, and distinguish the role of the markers by comparing model behavior before and after suppression. In detail, we design two complementary suppression methods that intervene at different levels of the generation process. \textbf{Prompt suppression} is a soft and generalizable intervention: it guides the model through a global natural-language instruction to avoid targeted anthropomorphic markers and adopt an objective style. 
\textbf{Token suppression} is a hard but less generalizable intervention: it directly blocks predefined target token sequences during decoding, offering precise lexical control but only over the registered markers.
Our results suggest that anthropomorphic markers are not uniformly necessary for problem-solving ability; beyond prior efficiency-oriented observations \cite{wang2025wait} on a few reflection keywords, we show that suppressing a broader set of markers can preserve or improve performance, especially under larger sampling budgets (\textit{RQ1}). These effects are specific to such markers (\textit{RQ2}), since suppressing other words leads to weaker performance. Our work further shows that markers are partially associated with explicit reflection (\textit{RQ3}), because models can still perform marker-free verification and revision after marker suppression. 
In terms of the mechanism of markers, they primarily alter the surface organization and sampling distribution of reasoning trajectories rather than the model's underlying reasoning ability (\textit{RQ4}). These could motivate future work to investigate internal reasoning mechanisms.


Our contributions are as follows:
\begin{itemize}
    \item We revisit the mechanism of anthropomorphic reflection markers, which focus on their necessity for reasoning and role in the reflection.
    \item We introduce prompt suppression and token suppression as complementary interventions for comprehensive analysis.
    \item Experiments show that such marks may not be uniformly necessary for reasoning and tend to be surface cues rather than reliable proxies for reflection. The model could also possibly turn to marker-free reflection as an alternative.
\end{itemize}

\section{Related Works}
\subsection{LLM Reasoning}

Recent work increasingly views reasoning as an inference-time process, where models improve performance by generating longer traces, sampling multiple trajectories, or selecting among candidate solutions~\citep{han2025token,xu2025chain,liu2025thought,muennighoff2025s1,brown2024large,snell2025scaling}. 
DeepSeek-R1 further shows that reinforcement learning can elicit long reasoning traces with self-reflection, verification, and strategy adaptation~\citep{guo2025deepseek,chen2024not,wang2025wait}. 
These studies highlight the importance of trajectory exploration and selection, but they do not clarify which visible trace components are actually necessary for reasoning. 
In particular, the mechanism of anthropomorphic reflection markers remains unclear.

\subsection{Self-Reflection and Overthinking}

Recent work has questioned whether visible reasoning faithfully reflects internal reasoning. Quiet-STaR \citep{zelikman2024quiet} and studies on CoT faithfulness \citep{paul2024making,tanneru2024hardness} suggest that useful reasoning may occur without being fully exposed in generated rationales, motivating a distinction between visible reflection and internal verification or search.
At the same time, long CoT can lead to overthinking and inefficient generation \citep{chen2024not,team2025kimi}. NoWait \citep{wang2025wait} shows that suppressing reflection keywords such as \textit{Wait} and \textit{Hmm} can shorten reasoning while preserving or improving performance, and studies of aha moments \citep{yang2025understanding} associate anthropomorphic tones with self-reflection and uncertainty adjustment. In contrast, unlike NoWait~\citep{wang2025wait}, which focuses on improving reasoning efficiency by suppressing a few reflection keywords, our work treats suppression as a probe: we study broader anthropomorphic markers, compare soft prompt suppression with hard token suppression to analyze whether anthropomorphic markers are necessary for reasoning, whether their effects are marker-specific, and whether reflection will turn into marker-free forms. We further analyze Pass@$k$ to reveal how markers affect the sampling distribution of reasoning trajectories for a more comprehensive view.

\section{Method}

\begin{table}[t]
\centering
\small
\begin{tabular}{c}
\hline
\textbf{Anthropomorphic Markers} \\
\hline
Aha, I, Hmm, Oh, Umm,  
Well, alright,  we,  Wait, \\ Alternatively, 
Case, Okay, perhaps, Still, we, 
We, \\ Let, Alright, Good, Consider, 
Adjust, Attempt,\\ Notice, Try, Test, 
Using, Oh, Oops, Break, \\ Looking,   
Attempting, Again, Like, Assume, Maybe  \\\hline
\end{tabular}
\caption{List of anthropomorphic markers.}
\label{tab:words}
\end{table}

We study anthropomorphic reflection markers as observable handles for probing explicit reflection, rather than as isolated lexical items. 
Since explicit reflection is often introduced or signaled by such markers, suppressing them allows us to test whether reasoning performance and reflection-like behavior depend on marker-associated reflection. 
Following \citet{yang2025understanding}, we define anthropomorphic markers as words that frequently appear at the beginning of reflective sentences and are more frequent in reflective responses than in non-reflective ones. 
Table~\ref{tab:words} lists the markers used in this study, and we suppress both capitalized and lower-case forms.
Specifically, we design two suppression methods: \textbf{Prompt suppression} and \textbf{Token suppression}.

\subsection{Prompt Suppression}

Prompt suppression adds an instruction requiring the model to reason step by step using formal and objective language while avoiding the target expressions. This is a shallow and high-level intervention. It does not directly forbid tokens, but it changes the model's expected output style and may reshape the entire reasoning trajectory. Its purpose is to examine how global stylistic and trajectory-level constraints affect visible reflection and reasoning performance. The example of the prompt included a statement is shown in Figure~\ref{fig:prompt}.

\begin{figure}[t]
\centering
\begin{tcolorbox}[
  colback=blue!5,
  colframe=blue!60!black,
  title=Prompt Suppression
]
\textbf{User:} Henry and 3 of his friends order 7 pizzas for lunch. Each pizza is cut into 8 slices. If Henry and his friends want to share the pizzas equally, how many slices can each of them have?

Please reason step by step, using formal, objective language only. Do not use informal or conversational expressions, including:
`aha', `Aha', `i', `I', `hmm', `Hmm', `oh', `Oh', \ldots, `maybe', `Maybe'.
Put your final answer within \texttt{\textbackslash boxed\{\}}.

\textbf{Assistant:}
\end{tcolorbox}
\caption{An example prompt used for prompt suppression.}
\label{fig:prompt}
\end{figure}

\subsection{Token Suppression}

Token suppression directly masks target token sequences during decoding. 
Different from prompt suppression, which constrains the model through natural-language instructions, token suppression intervenes locally on the decoding distribution. 
It therefore provides a lower-level probe for testing whether blocking anthropomorphic reflection markers prevents reflection-like behavior or harms task-solving ability without changing the overall prompt format.

Let $\mathcal{V}$ denote the vocabulary, and let $\mathbf{y}_{<t} = (y_1,\ldots,y_{t-1})$ be the generated prefix at decoding step $t$. 
Given the model logits $\mathbf{z}_t \in \mathbb{R}^{|\mathcal{V}|}$, the original next-token distribution is
\begin{equation}
p_\theta(y_t = v \mid \mathbf{x}, \mathbf{y}_{<t})
=
\frac{\exp(z_{t,v})}
{\sum_{u \in \mathcal{V}} \exp(z_{t,u})},
\end{equation}
where $\mathbf{x}$ is the input prompt and $v \in \mathcal{V}$ is a candidate next token.

We define a set of banned marker sequences as
\begin{equation}
\mathcal{S} = \{ \mathbf{s}^{(j)} \}_{j=1}^{J},
\quad
\mathbf{s}^{(j)} = (s^{(j)}_1, \ldots, s^{(j)}_{L_j}),
\end{equation}
where each $\mathbf{s}^{(j)}$ is the tokenized form of a target anthropomorphic marker, and $L_j$ is its token length. 
At each decoding step, we identify the set of tokens that would complete a banned marker sequence given the current prefix:
\begin{equation}
\begin{aligned}
\mathcal{B}_t
=
\{\, s^{(j)}_{L_j}
\mid\;&
\mathbf{s}^{(j)} \in \mathcal{S}, \\
&
\mathbf{y}_{t-L_j+1:t-1}
=
\mathbf{s}^{(j)}_{<L_j}
\,\}.
\end{aligned}
\end{equation}
Here, $\mathbf{s}^{(j)}_{<L_j}$ denotes the prefix
$(s^{(j)}_1,\ldots,s^{(j)}_{L_j-1})$ and $\mathcal{B}_t$ contains the final tokens of banned sequences whose preceding tokens have already been generated. 
We then apply logit masking:
\begin{equation}
\tilde{z}_{t,v}
=
\begin{cases}
-\infty, & v \in \mathcal{B}_t, \\
z_{t,v}, & v \notin \mathcal{B}_t,
\end{cases}
\end{equation}
and sample from the modified distribution
\begin{equation}
\tilde{p}_\theta(y_t = v \mid \mathbf{x}, \mathbf{y}_{<t})
=
\frac{\exp(\tilde{z}_{t,v})}
{\sum_{u \in \mathcal{V}} \exp(\tilde{z}_{t,u})}.
\end{equation}
This makes the probability of generating any token in $\mathcal{B}_t$ equal to zero, thereby preventing the model from completing the corresponding marker sequence.
For example, if the target marker \textit{Alternatively} is tokenized as
$\mathbf{s} = [201, 305, 412]$, the sequence is registered in $\mathcal{S}$.
During generation, once the prefix ends with $[201,305]$, the next-token candidate $412$ is added to $\mathcal{B}_t$ and its logit is set to $-\infty$.
As a result, the complete marker \textit{Alternatively} cannot be generated, while other tokens remain available.
This design blocks marker realization at the lexical level while leaving the rest of the decoding process unchanged.



\section{Experimental Setup}

\subsection{Datasets}
We evaluate our method on BIG-Bench Hard \citep{suzgun2023challenging}, MMLU-Pro \citep{wang2024mmlu}, AIME 2024 \citep{maa_aime}, and GSM8K \citep{cobbe2021training}. BIG-Bench Hard \citep{suzgun2023challenging} and MMLU-Pro \citep{wang2024mmlu} evaluate broad reasoning abilities across hard reasoning tasks and academic domains. AIME 2024 \citep{maa_aime} and GSM8K \citep{cobbe2021training} focus on mathematical reasoning, ranging from challenging high-school competition problems to grade-school arithmetic word problems. In our experiments, we use all $6,510$ BIG-Bench Hard test examples, all $12,032$ MMLU-Pro test examples, all $30$ AIME 2024 problems, and all $1,319$ GSM8K test problems.

\subsection{Compared Baseline}
We compare our suppression-based settings with a \textbf{No Suppression} baseline, where the original model is used without suppressing any reflection-related words. This setting measures the model's standard reasoning behavior and default task performance.

\subsection{Evaluation Metrics}
We evaluate task-solving ability using exact-match accuracy on the test data. Since a single sampled response may not fully reflect the model's reasoning capability, we report Pass@$k$ with $k \in \{1,2,8,16\}$ following prior work on repeated sampling and test-time scaling~\citep{brown2024large,snell2025scaling}. Pass@$1$ measures the correctness of the first sampled trajectory and reflects the model's default response quality. However, Pass@$1$ can be sensitive to sampling variance and may underestimate the model's ability when the first trajectory is incorrect but other valid reasoning paths are still accessible. Therefore, we additionally report Pass@$2$, Pass@$8$, and Pass@$16$ to evaluate whether a correct solution can be recovered within a finite sampling budget. These metrics provide a more comprehensive view of the model's problem-solving potential beyond its first response.

\subsection{Implementation Details}
We use DeepSeek-R1-Distill-Qwen-$1.5$B and DeepSeek-R1-Distill-Qwen-$7$B \citep{guo2025deepseek,yang2024qwen25math} as the backbone reasoning models. We choose these two models because they are representative open-source R1-style distilled models with different scales, allowing us to examine whether reflection-related behaviors are consistent across model sizes. Their explicit long-form reasoning traces also make them suitable for analyzing the role of anthropomorphic marker-based reflection in model reasoning.
We use the following decoding hyperparameters: temperature $=0.6$, top-$p=0.95$, and maximum generation length $=4096$ tokens. For each experimental setting, we conduct five independent runs and report the average results. All the experiments are conducted on A$100$ GPUs.

\begin{table*}[t]
\centering
\small
\resizebox{\textwidth}{!}{
\begin{tabular}{l|l|cccc|cccc}
\toprule
\multirow{2}{*}{\textbf{Model}} & \multirow{2}{*}{\textbf{Setting}}
& \multicolumn{4}{c}{\textbf{BIG-Bench Hard}}
& \multicolumn{4}{c}{\textbf{MMLU-Pro}} \\
\cmidrule(lr){3-6}
\cmidrule(lr){7-10}
& & \textbf{$k=1$} & \textbf{$k=2$} & \textbf{$k=8$} & \textbf{$k=16$}& \textbf{$k=1$} & \textbf{$k=2$} & \textbf{$k=8$} & \textbf{$k=16$} \\
\midrule

\multirow{3}{*}{\rotatebox{0}{1.5B}}
& No Suppression   & 28.0 & 36.7 & 56.0 & 64.5 & 7.7 & 14.2 & 29.6 & 37.8  \\
& Prompt Suppression & \textbf{33.1} & \textbf{44.3} & \textbf{69.4} & \textbf{81.3} & \textbf{9.4} & \textbf{18.4} &\textbf{33.8} & \textbf{43.4}  \\
& Token Suppression  & 28.5 & 36.1 & 55.1& 64.2  & 8.9 & 15.5 & 30.2 & 38.8  \\

\midrule

\multirow{3}{*}{\rotatebox{0}{7B}}
& No Suppression     & 63.7 & \textbf{78.1} &88.6& 90.9  & 38.7 & 48.1 & 64,3 & 70.1  \\
& Prompt Suppression & 61.6 & 75.8 & 89.0&91.1  & 32.7 & 43.6 & 63.4 & 70.2  \\
& Token Suppression  & \textbf{66.0} & \textbf{78.1} & \textbf{89.2}& \textbf{91.5}  & \textbf{41.5}& \textbf{50.0} & \textbf{65.7} & \textbf{72.0} \\

\bottomrule
\end{tabular}
}
\caption{Evaluation results on different general benchmarks. We report the Pass@$k$ with $k \in \{1,2,8,16\}$ results for a comprehensive analysis. The \textbf{bold} denotes the best results.}
\label{tab:main_general}
\end{table*}

\begin{table*}[t]
\centering
\small
\resizebox{\textwidth}{!}{
\begin{tabular}{l|l|cccc|cccc}
\toprule
\multirow{2}{*}{\textbf{Model}} & \multirow{2}{*}{\textbf{Setting}}
& \multicolumn{4}{c}{\textbf{AIME 2024}}
& \multicolumn{4}{c}{\textbf{GSM8K}} \\
\cmidrule(lr){3-6}
\cmidrule(lr){7-10}
& & \textbf{$k=1$} & \textbf{$k=2$} & \textbf{$k=8$} & \textbf{$k=16$} & \textbf{$k=1$} & \textbf{$k=2$} & \textbf{$k=8$} & \textbf{$k=16$} \\
\midrule

\multirow{3}{*}{\rotatebox{0}{1.5B}}
& No Suppression     & \textbf{15.1} & 20.7 & \textbf{30.3} & 35.3 & \textbf{78.6} & 86.8 & 94.1 & 95.9  \\
& Prompt Suppression & 6.7 & \textbf{23.3} & 30.0 & \textbf{40.0} & 78.2 & \textbf{87.1} & \textbf{94.3} & \textbf{96.0}   \\
& Token Suppression  & 10.1 & 15.6 & 28.6& 36.3  & 75.6 & 84.6 & 93.1 & 95.1  \\

\midrule

\multirow{3}{*}{\rotatebox{0}{7B}}
& No Suppression    & 21.6 & 28.6 &42.7& 48.1  & \textbf{91.2} & \textbf{94.1} & 96.7 & 97.3  \\
& Prompt Suppression & 23.3 & 30.0 &40.0&46.7 & 87.7 & 92.7 & 96.3 & 98.0  \\
& Token Suppression  & \textbf{23.4} & \textbf{32.0} &\textbf{50.3}& \textbf{58.8}  & 89.5 & 93.8 & \textbf{97.7} & \textbf{98.3}  \\

\bottomrule
\end{tabular}
}
\caption{Evaluation results on different math benchmarks.  Meaning of the table marker is the same as in Table \ref{tab:main_general}.}
\label{tab:main_math}
\end{table*}

\section{Experimental Results and Analysis}
We now examine the mechanism of anthropomorphic reflection markers, which focus on their necessity for reasoning and role in the reflection. 
Accordingly, our analysis follows the four \textit{RQs} introduced in Section~\ref{sec:introduction}.
The following experiments and analyses are organized around these \textit{RQs}.

\subsection{\textit{RQ1}: Reflection and Problem-Solving Ability}
We report task performance w/ and w/o suppression in Tables \ref{tab:main_general} and \ref{tab:main_math}. Specifically, we observe these phenomena:

\paragraph{Suppressing anthropomorphic reflection markers often improves performance.} Suppression does not simply degrade task performance. In most settings, suppression improves performance, especially when evaluated with larger pass@$k$. For example, on BIG-Bench Hard with the $1.5$B model, Prompt Suppression improves Pass@$1$ from $28.0$ to $33.1$, and Pass@$16$ from $64.5$ to $81.3$. This suggests that such marks is not uniformly necessary for successful reasoning.

\paragraph{The benefit of suppression becomes more visible at larger Pass@$k$.} While Pass@$1$ measures the first sampled trajectory, larger $k$ values reflect whether correct solutions are recoverable within a finite sampling budget. Therefore, even if a suppression method fails to improve pass@$1$, it can still improve pass@$8$ or pass@$16$. In the Table \ref{tab:main_math}, on GSM8K with the $7$B model, Token Suppression fails to improve Pass@$1$ from $91.2$ to $89.5$, but substantially improves Pass@$8$ from $96.7$ to $97.7$ and Pass@$16$ from $97.3$ to $98.3$. This suggests that suppression does not necessarily make the first response much better, but can make correct reasoning paths more accessible and enlarge the reasoning potential across multiple samples.

\paragraph{The trend is more consistent on general reasoning benchmarks than on mathematical benchmarks.} On BIG-Bench Hard and MMLU-Pro as shown in the Table~\ref{tab:main_general}, suppression consistently improves performance, especially for the $1.5$B model under Prompt Suppression and the $7$B model under Token Suppression. In contrast, math benchmarks show more unstable behavior as shown in the Table~\ref{tab:main_math}. For example, on GSM8K with the $7$B model, No Suppression achieves the best Pass@$1$ and Pass@$2$ scores (i.e., $91.2$ and $94.1$), while Token Suppression performs best at Pass@$8$ and Pass@$16$ ($97.7$ and $98.3$). This illustrates that general tasks tend to show more consistent trends because they often allow multiple compact reasoning strategies. In these settings, style regularization and reduced verbosity can help without strongly disrupting the solution process. Mathematical tasks are less stable because they depend more on precise intermediate computation, stepwise verification, and backtracking. Suppression may remove redundant reflection in some cases, but it may also interfere with useful deliberation in others.

\paragraph{The results reveal a scale-dependent pattern.} The $1.5$B model benefits more from Prompt Suppression, while the $7$B model benefits more from Token Suppression. For example, on BIG-Bench Hard with the $1.5$B model as shown in the Table~\ref{tab:main_general}, Prompt Suppression improves Pass@16 from $64.5$ to $81.3$, whereas Token Suppression slightly decreases it to $64.2$. This suggests that smaller models may have weaker understanding of instructions and flexibility, making that token-level suppression may disrupt generation in ways that the model cannot reliably compensate for. In contrast, the $7$B model shows strong gains under Token Suppression, such as on MMLU-Pro where Pass@$16$ improves from $70.1$ to $72.0$ as shown in the Table~\ref{tab:main_general}, and on AIME 2024 where it improves from $48.1$ to $58.8$ as shown in the Table~\ref{tab:main_math}. This indicates that larger models have stronger instruction-following ability and richer paraphrastic alternatives. It can avoid banned tokens while preserving verification or revision behavior through other expressions, making token suppression less damaging and often more effective.

\begin{table}[t]
\centering
\resizebox{0.5\textwidth}{!}{
\begin{tabular}{lcccc}
\toprule
\textbf{Setting} & \textbf{BBH} & \textbf{MM.} & \textbf{AIME.} & \textbf{GSM.} \\

\midrule
\rowcolor{blue!5} \multicolumn{5}{c}{\textit{DeepSeek-R1-Distill-Qwen-$1.5$B}}\\ \midrule

\textbf{Prompt Suppression} & \textbf{33.1} & \textbf{9.4} & \textbf{6.7} & \textbf{78.2}  \\
 \ \  w/ random markers & 32.4 & 7.8 & 6.5 & 77.8 \\
 \ \  w/ Fre. markers & 32.2 & 7.9 & 6.5 & 78.0 \\
\cmidrule(lr){1-5}
\textbf{Token Suppression}  & \textbf{28.5} & \textbf{8.9} & \textbf{10.1} & \textbf{75.6}  \\
 \ \  w/ random markers & 25.7 & 8.8 & 7.3 & 74.6 \\
 \ \  w/ Fre. markers & 27.5 & 8.2 & 10.0 & 74.4 \\

\midrule

\rowcolor{blue!5} \multicolumn{5}{c}{\textit{DeepSeek-R1-Distill-Qwen-$7$B}}\\ \midrule

\textbf{Prompt Suppression} & \textbf{61.6} & \textbf{32.7} & \textbf{23.3} & \textbf{87.7}  \\
 \ \  w/ random markers & 59.2 & 30.9 & 20.2 & 87.6 \\
 \ \  w/ Fre. markers & 58.9 & 31.7 & 16.1 & 86.9 \\
\cmidrule(lr){1-5}
\textbf{Token Suppression}  & \textbf{66.0} & \textbf{41.5} & \textbf{23.4} & \textbf{89.5}  \\
 \ \  w/ random markers & 61.5 & 39.0 & 20.1 & 88.8 \\
 \ \  w/ Fre. markers & 64.9 & 39.4 & 20.5 & 88.4 \\

\bottomrule
\end{tabular}
}
\caption{Ablation study on our suppression methods. We report the Pass@1 results. Note that BIG-Bench Hard denotes BIG-Bench Hard, MM. denotes MMLU-Pro, AIME. denotes AIME 2024 and GSM. denotes GSM8K. The \textbf{bold} denotes the best results.}
\label{tab:ablation}
\end{table}

\subsection{\textit{RQ2}: Specificity to Reflection Markers}
To examine whether the observed effects are specific to anthropomorphic reflection markers, we compare our original suppression setting with two control variants: \textit{random markers} and \textit{frequency-matched markers}. Random markers are randomly selected tokens, while frequency-matched markers are selected to match the word frequency of anthropomorphic markers under each setting. The specific markers we used in the ablation study are provided in the Appendix~\ref{ap:marker}. As shown in Table~\ref{tab:ablation}, suppressing anthropomorphic markers generally achieves better Pass@$1$ than both control variants, suggesting that the effect is not merely caused by removing arbitrary tokens or frequent words.

For the $1.5$B model, Prompt Suppression with anthropomorphic markers achieves the best performance on all benchmarks, reaching $33.1$ on BIG-Bench Hard, $9.4$ on MMLU-Pro, $6.7$ on AIME 2024, and $78.2$ on GSM8K. In contrast, random-marker suppression and frequency-matched suppression consistently perform worse, such as on MMLU-Pro where performance drops from $9.4$ to $7.8$ and $7.9$, respectively. A similar trend appears under Token Suppression: suppressing anthropomorphic markers obtains $28.5$ on BIG-Bench Hard and $10.1$ on AIME 2024, outperforming random markers (i.e., $25.7$ and $7.3$) and frequency-matched markers (i.e., $27.5$ and $10.0$).
The same pattern is more evident for the $7$B model. Under Prompt Suppression, anthropomorphic-marker suppression achieves $61.6$ on BIG-Bench Hard and $23.3$ on AIME 2024, while random-marker suppression drops to $59.2$ and $20.2$, and frequency-matched suppression further drops to $58.9$ and $16.1$. Under Token Suppression, the original setting also consistently performs best, reaching $66.0$ on BIG-Bench Hard, $41.5$ on MMLU-Pro, $23.4$ on AIME 2024, and $89.5$ on GSM8K. By comparison, random-marker suppression decreases these scores to $61.5$, $39.0$, $20.1$, and $88.8$, respectively.

These results indicate that the improvements are not simply due to generic lexical disruption or the suppression of high-frequency words. Instead, anthropomorphic reflection markers appear to occupy a specific functional role in organizing visible reflection behavior. Suppressing them affects the model differently from suppressing random or frequency-matched tokens, supporting the view that the observed performance changes are specifically tied to marker-associated reflection rather than arbitrary token removal.

\subsection{\textit{RQ3}: Reflection-Marker Correlation}

\begin{figure*}
    \centering
    \includegraphics[width=0.99\linewidth]{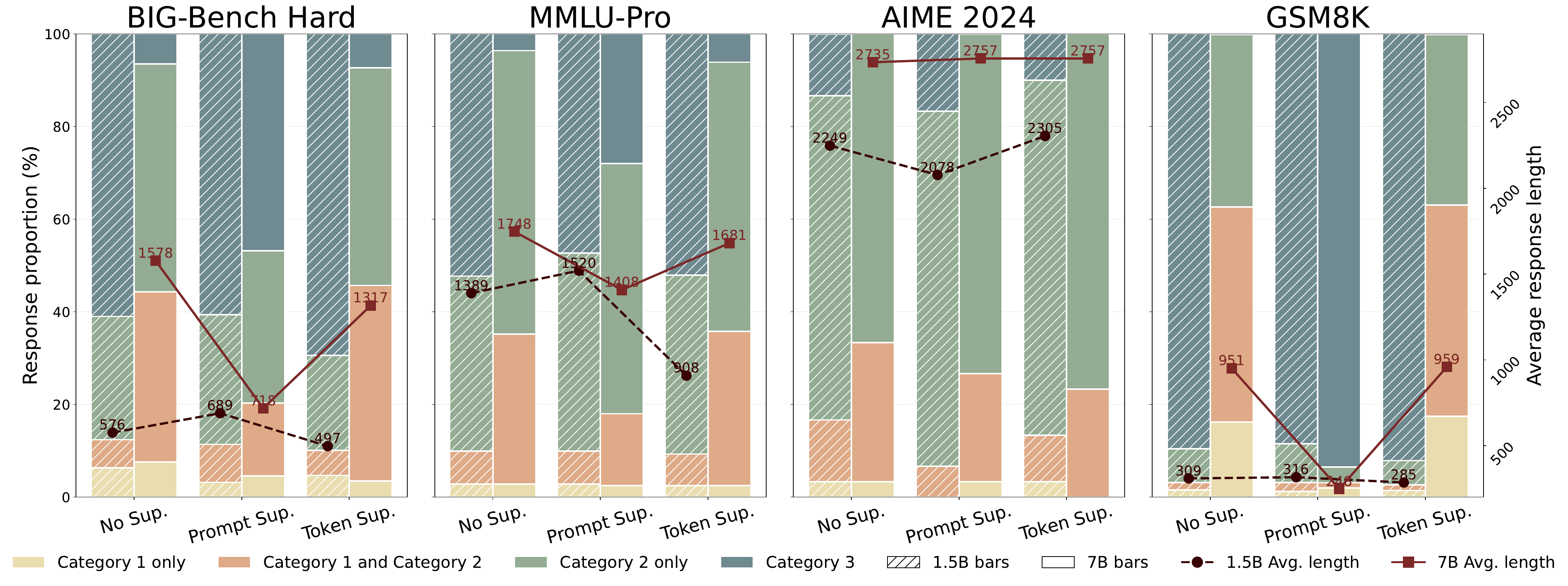}
    \caption{Distribution of explicit reflection behaviors and average response lengths.}
    \label{fig:judge}
\end{figure*}

\begin{figure}
    \centering
    \includegraphics[width=0.99\linewidth]{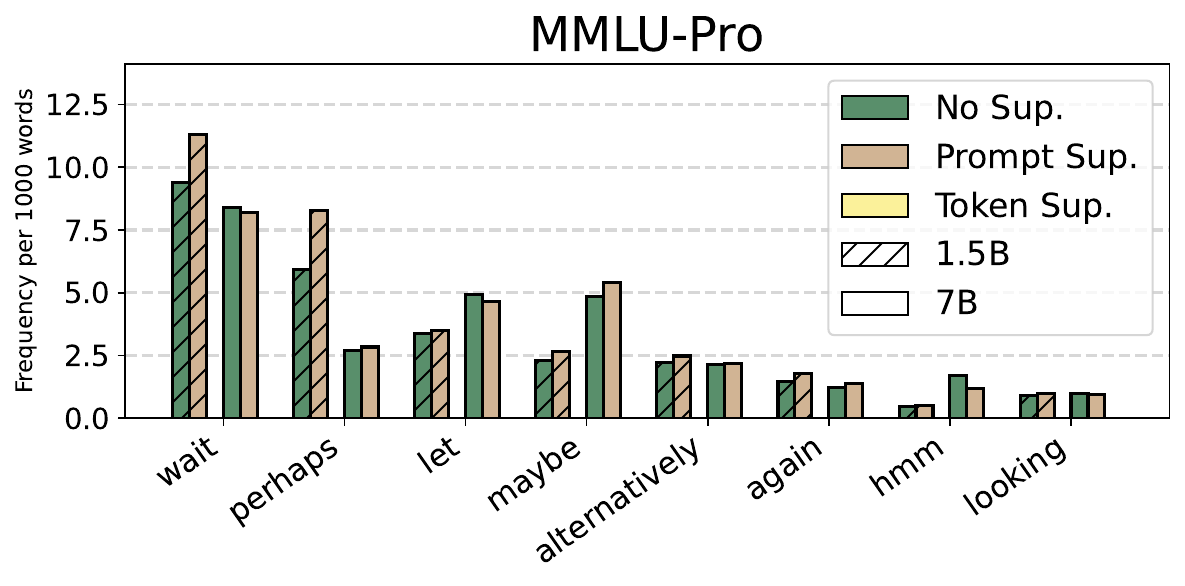}
    \caption{Frequency of anthropomorphic markers in the response of MMLU-Pro. }
    \label{fig:word_fre_mmlu}
\end{figure}

To analyze whether reflection still exists after suppressing the anthropomorphic markers, following recent work that uses strong LLMs as scalable evaluators~\citep{zheng2023judging,liu2023geval}, we use GPT-5.4-mini \citep{singh2025openai} as external LLM judge to classify model responses into three reflection categories: (1) Category 1 denotes \textit{solution followed by reflection}, where the model first derives an answer and then verifies or reconsiders it. (2) Category 2 denotes \textit{reflection while reasoning}, where the model performs verification before reaching the final answer. (3) Category 3 denotes \textit{no reflection}, where the response contains no reasoning-level verification or alternative reasoning. Note that we also allow multiple reflection categories to appear, only when the response is assigned to both Category 1 and Category 2. The detail setting of the external LLM judge is provided in the Appendix~\ref{ap:judge}.

Results on all the suppression settings are shown in Figure~\ref{fig:judge}, \textbf{suppressing anthropomorphic markers does not necessarily remove explicit reflection behavior}. Note that we also analyze the reflection category shift as shown in Appendix~\ref{ap:rq3_shift}. In many settings, especially for the $7$B model, responses under Token Suppression still contain a large proportion of Category 2 or Category 1$+$2. For example, on BIG-Bench Hard and MMLU-Pro, the $7$B model under Token Suppression preserves substantial reflection-related categories, suggesting that reflection can take multiple forms (e.g., marker-free) and remain explicit through alternative expressions. Anthropomorphic markers are the most explicit form, but suppressing this marker-associated reflection does not fully remove reflection behavior. Models can still verify their reasoning by bypassing the suppressed words or using alternative expressions. Moreover, reflection is associated with longer responses, suggesting that it plays a functional role in extended reasoning chains rather than merely serving as a stylistic marker.


\begin{figure*}[th]
    \centering
    \includegraphics[width=0.99\linewidth]{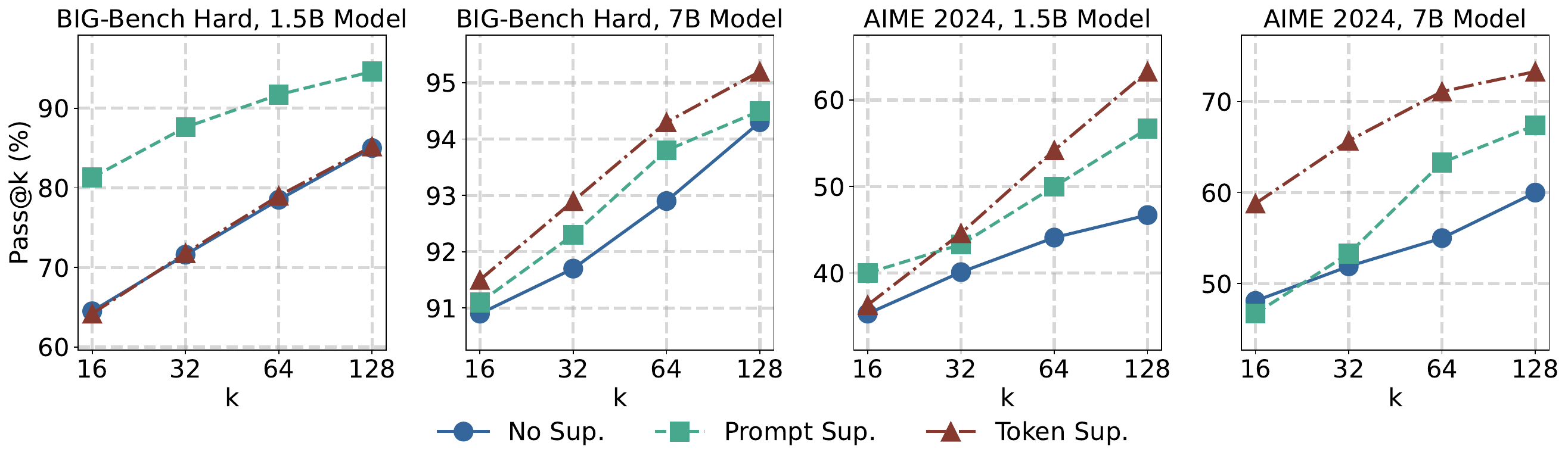}
    \caption{Pass@$k$ results for larger $k$.}
    \label{fig:passk}
\end{figure*}

\begin{figure}[th]
    \centering
    \small
    \begin{tcolorbox}[
  colback=blue!3,
  colframe=blue!50!black,
  coltitle=black,
  colbacktitle=blue!15,
  title=\textbf{Model Response}
]
    To solve this problem, it’s best to define variables for each type of tree’s
apples ... \\
\textcolor{red}{\textbf{There must be a mistake in my understanding. Re-reading the problem:}} ... \\
But the initial answer given by the assistant was 170, \textcolor{red}{\textbf{which was incorrect}} because it didn't account for the sister's apples.
So, the correct answer is 350.
    \end{tcolorbox}
    \caption{An example of model response from the Token Suppression condition in GSM8K.}
    \label{fig:case}
\end{figure}

We also notice that the stability of reflection behavior is related to model scale. Smaller models appear more sensitive to suppression, while larger models tend to preserve reflection behavior more robustly. For instance, on AIME 2024, Prompt Suppression reduces the proportion of reflection-related responses for the $1.5$B model, indicating that prompt-level constraints can suppress explicit reflection in smaller models. In contrast, for the $7$B model, all settings still contain reflection-related responses on AIME 2024, regardless of whether suppression is applied. This suggests that larger models are more inclined to perform reflection on complex tasks, likely because they can bypass lexical constraints and express verification or revision through alternative formulations. Therefore, these suggest that reflection, broadly defined to include both marker-associated and marker-free forms, should not be treated simply as equivalent to the presence of anthropomorphic markers; rather, it is influenced by model capacity and task complexity. \textbf{Larger models may rely more stably on reflection for difficult reasoning and can express it in more flexible forms}.


\subsection{\textit{RQ4}: Mechanistic Implications}

\paragraph{Word Frequency Shift.} To analyze whether suppression actually works, we further analyze the word frequency of anthropomorphic reflection markers in model responses. We measure marker frequency by counting the banned anthropomorphic markers in each model response. The visualization is shown in Figure~\ref{fig:word_fre_mmlu}, we focus on MMLU-Pro and plot the most frequent markers under the no-suppression setting. Suppression affects markers unevenly: For several high-frequency markers, such as \textit{wait}, Prompt Suppression can even increase the frequency relative to No Suppression, suggesting that simply instructing the model to avoid targeted anthropomorphic expressions does not reliably suppress the actual lexical markers used in reflection. These results illustrate that \textbf{marker and reflection are not fully associated.} Prompt Suppression may fail to suppress the targeted surface forms and can even induce alternative reflective expressions, while Token Suppression fully reduces banned markers while remaining reflection behavior itself. More results are provided in Appendix~\ref{ap:word_shift}.


\paragraph{Effect of Larger Search Space.} Figure~\ref{fig:passk} shows that suppression often surpasses the No Suppression baseline more clearly as $k$ increases. For example, on BIG-Bench Hard with the 1.5B model, Prompt Suppression remains consistently above the baseline and reaches nearly $95$ at $k=128$. These results suggest that suppression does not merely affect the first response, but increases the accessibility of correct reasoning trajectories under sampling.
This directly informs our central question. If anthropomorphic markers were necessary for reasoning, suppressing them should make correct solutions harder to recover as $k$ grows. Instead, the gains often become larger at higher $k$, indicating that models can still verify, search, and revise through marker-free reflection even when these markers are discouraged. Therefore, \textbf{anthropomorphic markers appear to change the surface organization and sampling distribution of reasoning trajectories}, rather than removing the model's underlying reasoning ability.

\subsection{Case Study}

Figure~\ref{fig:case} presents an example under Token Suppression. Even without anthropomorphic reflection markers, the model identifies a contradiction, re-reads the problem, and revises the solution. This shows that reflection can persist without explicit anthropomorphic markers, supporting our finding that such markers are not always necessary for reasoning and reflection.

\section{Conclusion}

In this work, we revisit anthropomorphic reflection markers in LLM reasoning through prompt and token suppression. 
Across four benchmarks and two model scales, we find that these markers are neither uniformly necessary nor merely ordinary words: suppressing them often preserves or improves performance, and differs from random or frequency-matched suppression. 
They are also only partial indicators of explicit reflection, as models can still verify and revise in marker-free forms. 
Overall, anthropomorphic markers mainly shape the surface organization and sampling distribution of reasoning trajectories, rather than the model's underlying reasoning ability.
Future work should develop more direct methods to investigate reflection mechanisms beyond observable markers.

\section*{Limitations}

This work has several limitations. First, our reflection analysis relies on an external LLM judge, which may introduce model-specific biases. Second, our experiments are limited to two DeepSeek-R1-Distill-Qwen models. The observed effects may vary for other model families or tasks requiring different types of verification. Finally, while our transition and case analyses suggest that reflection can be reformulated without anthropomorphic markers, they do not directly reveal the model’s internal reasoning mechanism. Future work should combine behavioral interventions with representation-level or activation-level analyses to more directly examine how reflection is implemented inside reasoning models.



\bibliography{main}

\appendix

\section{Suppressed Markers in \textit{RQ2}}
\label{ap:marker}

We provide the suppressed \textit{random markers} in the Table~\ref{tab:random_marker} and \textit{frequency-matched markers} in the Tables~\ref{tab:fre_marker_15_bbh},~\ref{tab:fre_marker_7_bbh},~\ref{tab:fre_marker_15_mm},~\ref{tab:fre_marker_7_mm},~\ref{tab:fre_marker_15_aime},~\ref{tab:fre_marker_7_aime},~\ref{tab:fre_marker_15_gsm},~\ref{tab:fre_marker_7_gsm}.

\begin{table}[t]
\centering

\begin{tabular}{>{\centering\arraybackslash}m{0.4\textwidth}}
\hline
\textbf{Anthropomorphic Markers} \\
\hline
near, large, small, early, late, simple, different, part,
kind, place, point, group, line, side, form, given,
shown, written, called, means, left, right, above, below,
first, second, next, last, total, value, result, possible,
certain, common, and whole  \\\hline
\end{tabular}
\caption{List of anthropomorphic markers suppressed in the \textit{random markers} setting.}
\label{tab:random_marker}
\end{table}

\begin{table}[t]
\centering

\begin{tabular}{>{\centering\arraybackslash}m{0.4\textwidth}}
\hline
\textbf{Anthropomorphic Markers} \\
\hline
shown, right, common, late, early, into, few, than, that,
problem, when, over, left, number, this, certain, another,
called, kind, large, expression, simple, many, near, place,
also, small, different, without, under, and above \\
\hline
\end{tabular}
\caption{List of anthropomorphic markers suppressed in the \textit{frequency-matched markers} setting for BIG-Bench Hard, $1.5B$ model.}
\label{tab:fre_marker_15_bbh}
\end{table}

\begin{table}[t]
\centering

\begin{tabular}{>{\centering\arraybackslash}m{0.4\textwidth}}
\hline
\textbf{Anthropomorphic Markers} \\
\hline
shown, which, when, early, late, over, those, problem,
that, question, both, also, another, number, then, common,
many, few, near, simple, last, small, place, kind, means,
either, called, than, given, certain, and from\\
\hline
\end{tabular}
\caption{List of anthropomorphic markers suppressed in the \textit{frequency-matched markers} setting for BIG-Bench Hard, $7B$ model.}
\label{tab:fre_marker_7_bbh}
\end{table}

\section{Details of the LLM-as-Judge Setting in \textit{RQ3}}
\label{ap:judge}

We provide the prompt used in the experiments in the Figures~\ref{fig:judge_prompt_1} and ~\ref{fig:judge_prompt_2}.

\section{Reflection Category Shift}
\label{ap:rq3_shift}
\begin{figure*}[h]
    \centering
    \includegraphics[width=0.99\linewidth]{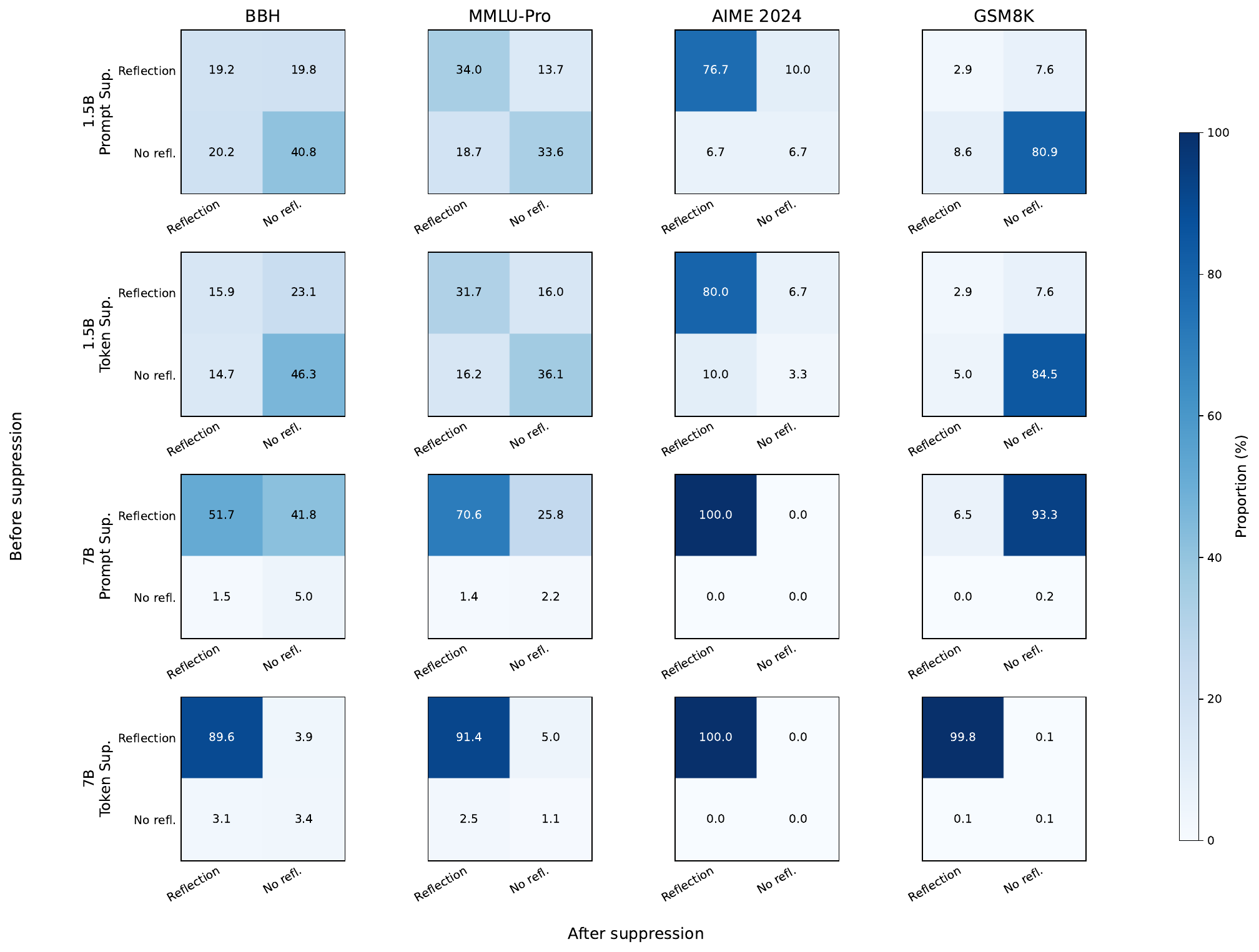}
    \caption{Transition matrices for $1.5$ and $7$B models on GSM8K, BIG-Bench Hard, MMLU-Pro, and AIME 2024.}
    \label{fig:rq3_shift}
\end{figure*}

The category distribution in \textit{RQ3} shows the overall frequency of reflection types, but not whether the same responses stop reflecting or shift between forms. We therefore analyze sample-level category transitions before and after suppression to further determine whether markers affect reflection or mainly change how it is expressed. Figure \ref{fig:rq3_shift} shows the full results. We just select two representative transition matrices for $7$B model on GSM8K for analysis, comparing prompt suppression and token suppression. Under prompt suppression, reflective responses are strongly shifted to No reflection. This indicates that prompt suppression can substantially reduce explicit reflection across the categories considered here. Token suppression shows the opposite pattern under the same model and benchmark. Almost no reflective responses move to No reflection. Instead, responses are mostly redistributed within reflective categories. Thus, token suppression does not remove reflection behavior here; it mainly changes the organization or surface realization of reflection. 

\section{More results on Word Frequency Shift}
\label{ap:word_shift}

\begin{figure}[h]
    \centering
    \includegraphics[width=0.99\linewidth]{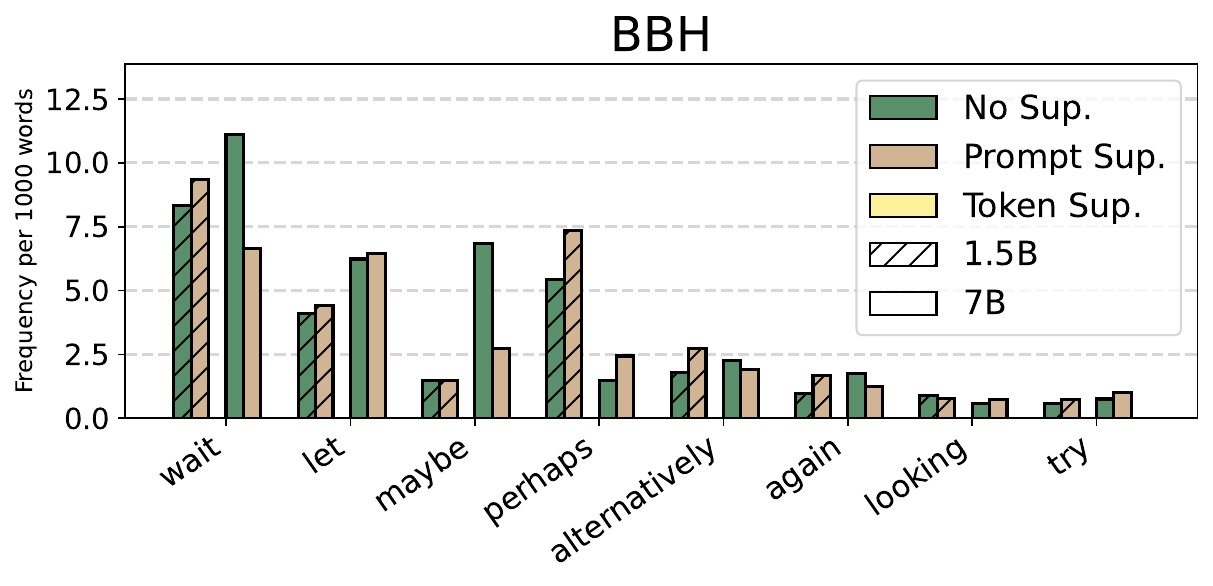}
    \caption{Word Frequency Shift on BIG-Bench Hard.}
    \label{fig:word_bbh}
\end{figure}

\begin{figure}[h]
    \centering
    \includegraphics[width=0.99\linewidth]{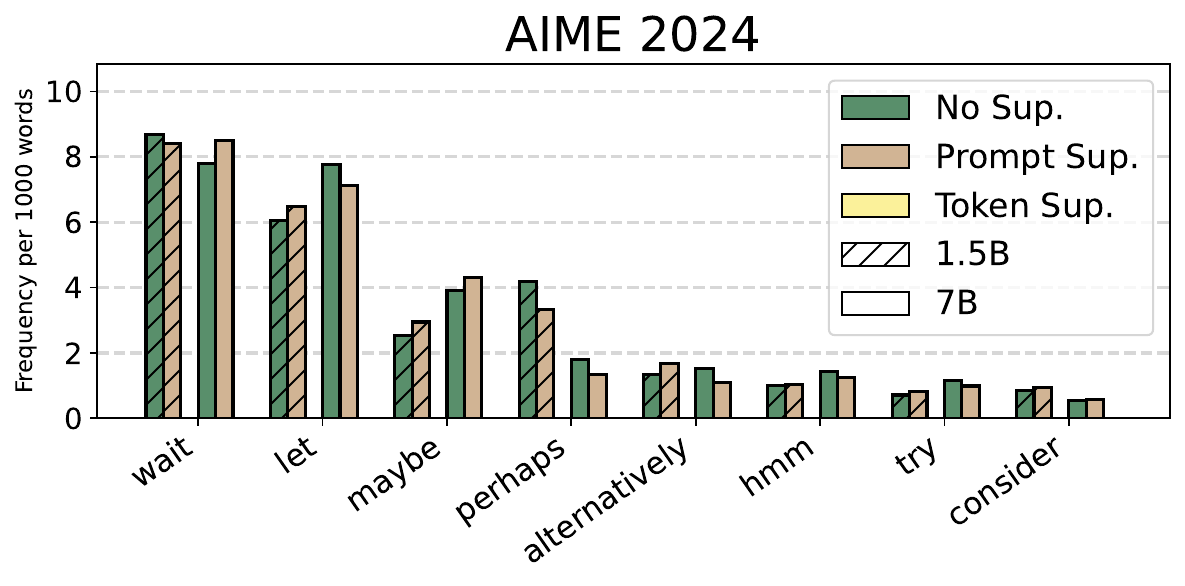}
    \caption{Word Frequency Shift on AIME 2024.}
    \label{fig:word_aime}
\end{figure}

\begin{figure}[h]
    \centering
    \includegraphics[width=0.99\linewidth]{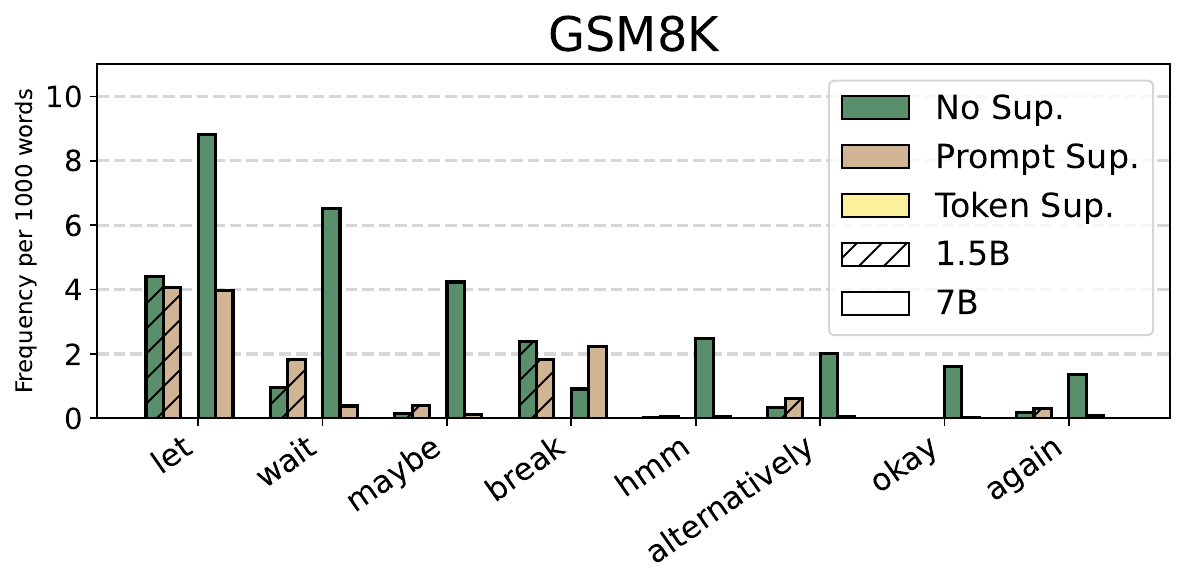}
    \caption{Word Frequency Shift on GSM8K.}
    \label{fig:word_gsm}
\end{figure}

Results on BIG-Bench Hard are shown in the Figure~\ref{fig:word_bbh}, AIME 2024 are shown in the Figure~\ref{fig:word_aime}, GSM8K are shown in the Figure~\ref{fig:word_gsm}.

\section{The Use of Large Language Models}

In the preparation of this manuscript, LLMs were used solely for the purpose of text polishing, including grammar correction and stylistic refinement. No content was generated or rewritten with the intention of altering the scientific meaning, originality, or conclusions of the work. All ideas, analyses, and results presented in this paper are entirely the authors' own.

\begin{table}[h]
\centering

\begin{tabular}{>{\centering\arraybackslash}m{0.4\textwidth}}
\hline
\textbf{Anthropomorphic Markers} \\
\hline
shown, that, how, common, late, small, near, from, which,
both, where, within, answer, less, question, large, step,
kind, called, early, without, group, when, last, expression,
first, place, each, another, simple, and given\\
\hline
\end{tabular}
\caption{List of anthropomorphic markers suppressed in the \textit{frequency-matched markers} setting for MMLU-Pro, $1.5B$ model.}
\label{tab:fre_marker_15_mm}
\end{table}

\begin{table}[h]
\centering

\begin{tabular}{>{\centering\arraybackslash}m{0.4\textwidth}}
\hline
\textbf{Anthropomorphic Markers} \\
\hline
shown, that, each, common, late, either, early, an, which,
both, given, these, from, without, because, line, possible,
kind, near, below, over, next, less, place, left, another,
few, total, different, called, and this\\
\hline
\end{tabular}
\caption{List of anthropomorphic markers suppressed in the \textit{frequency-matched markers} setting for MMLU-Pro, $7B$ model.}
\label{tab:fre_marker_7_mm}
\end{table}

\begin{table}[h]
\centering

\begin{tabular}{>{\centering\arraybackslash}m{0.4\textwidth}}
\hline
\textbf{Anthropomorphic Markers} \\
\hline
few, number, side, early, late, next, those, which, each,
given, total, more, from, large, this, within, step, near,
simple, common, over, under, small, question, written, after,
while, place, condition, result, and first\\
\hline
\end{tabular}
\caption{List of anthropomorphic markers suppressed in the \textit{frequency-matched markers} setting for AIME 2024, $1.5B$ model.}
\label{tab:fre_marker_15_aime}
\end{table}

\begin{table}[h]
\centering

\begin{tabular}{>{\centering\arraybackslash}m{0.4\textwidth}}
\hline
\textbf{Anthropomorphic Markers} \\
\hline
few, that, total, under, early, those, written, which, number,
same, first, either, these, many, from, called, part, late,
shown, near, also, place, possible, while, certain, without,
simple, form, over, within, and then\\
\hline
\end{tabular}
\caption{List of anthropomorphic markers suppressed in the \textit{frequency-matched markers} setting for AIME 2024, $7B$ model.}
\label{tab:fre_marker_7_aime}
\end{table}

\begin{table}[h]
\centering

\begin{tabular}{>{\centering\arraybackslash}m{0.4\textwidth}}
\hline
\textbf{Anthropomorphic Markers} \\
\hline
near, left, either, kind, shown, within, form, each, an,
without, question, few, small, also, first, large, less, late,
simple, called, expression, other, between, below, problem,
common, term, part, right, early, and different\\
\hline
\end{tabular}
\caption{List of anthropomorphic markers suppressed in the \textit{frequency-matched markers} setting for GSM8K, $1.5B$ model.}
\label{tab:fre_marker_15_gsm}
\end{table}

\begin{table}[h!]
\centering

\begin{tabular}{>{\centering\arraybackslash}m{0.4\textwidth}}
\hline
\textbf{Anthropomorphic Markers} \\
\hline
shown, total, many, kind, near, group, where, this, then,
after, part, left, what, less, each, value, without, expression,
condition, late, before, simple, between, few, question, form,
called, because, result, point, and first\\
\hline
\end{tabular}
\caption{List of anthropomorphic markers suppressed in the \textit{frequency-matched markers} setting for GSM8K, $7B$ model.}
\label{tab:fre_marker_7_gsm}
\end{table}

\begin{figure*}[th]
    \centering
    \small
\begin{tcolorbox}[
  colback=blue!5,
  colframe=blue!60!black,
  title=Prompt of the External LLM Judge (Part One)
]
You are an evaluator. Your task is to classify the response into one or more of the following categories.

In this task, ``reflection'' refers to verification- or revision-oriented reasoning behavior. It includes actions such as checking previous steps, recalculating, testing intermediate claims with concrete cases, reconsidering assumptions, correcting mistakes, or trying an alternative reasoning path to verify whether the same final answer can be derived.

\paragraph{Category 1: Solution followed by reflection.}
The model first performs reasoning and derives a final answer. After the final answer is obtained, the model performs reflection to verify, reconsider, or revise the reasoning that led to the final answer. This reflection targets the correctness and internal consistency of the reasoning and may involve recalculating, reviewing previous steps, considering simpler conditions to test whether the reasoning holds, or applying an alternative reasoning approach to determine whether the same final answer can be derived. Expressions that merely state confidence, reasonability, or uncertainty after the final answer are not sufficient for this category unless they include reasoning-level verification, reconsideration, correction, or alternative reasoning.

Here are two illustrative examples:

Therefore, $m = 25$ and $n = 8$, so $m + n = 33$.

Wait, let me just double-check my steps because I want to make sure I did not make a mistake.

Starting from the beginning, the logs were converted correctly to exponents. Then, substitution steps were followed, and each substitution seemed correct. Found $z$ first, \ldots

So, the number of such paths is $\boxed{294}$.

But, just to be thorough, is there another way to think about this?

Another approach is to model the number of direction changes.

These examples illustrate the pattern where the model states a final answer first, such as ``$m+n=33$'' or ``$\boxed{294}$'', and then performs reflection after the final answer is obtained. This shows that reflection is carried out after the final answer to verify or reconsider the correctness of the reasoning, either by reviewing the previous reasoning or by applying an alternative reasoning approach. These examples are illustrative and do not define all possible cases.

\paragraph{Category 2: Reflection while reasoning.}
The model performs reflection during reasoning, before a final answer is derived. This reflection is applied to intermediate parts of the reasoning and is used to verify, reconsider, correct, or update the ongoing reasoning. As a result of reflection, the reasoning may be revised and updated in order to proceed toward a valid final answer. Expressions that merely state confidence, uncertainty, reasonability, or preference during reasoning are not sufficient for this category unless they include reasoning-level verification, reconsideration, correction, or alternative reasoning.

Here are two illustrative examples:

If a path has four direction changes, how many segments of right and up moves does it have? Each direction change corresponds to a switch from right to up or up to right, so the number of segments is one more than the number of direction changes. So, four direction changes would result in five segments.

Wait, let's test that. Suppose I start with a right move. Then, if I change direction four times, I would have right, up, right, up, right. That's four direction changes and five segments. Alternatively, starting with an up move would give up, right, up, right, up, which is also four direction changes and five segments. So, yes, four direction changes lead to five segments.

So now I have three equations:
\begin{enumerate}
    \item $a - b - c = \frac{1}{2}$
    \item $-a + b - c = \frac{1}{3}$
    \item $-a - b + c = \frac{1}{4}$
\end{enumerate}

Wait, hold on, let me write them correctly:

Equation (1): $a - b - c = \frac{1}{2}$

Equation (2): $-a + b - c = \frac{1}{3}$

Equation (3): $-a - b + c = \frac{1}{4}$

Let me try elimination.

\ldots

Wait, that's interesting. $c$ is negative, but since $z$ is a positive real number, $\log_2 z$ can be negative, meaning $z$ is less than $1$. So that's okay.

\ldots

Which matches, so that's correct.

\end{tcolorbox}
\caption{An example prompt used for external LLM judge (Part One).}
    \label{fig:judge_prompt_1}
\end{figure*}

\begin{figure*}[th]
    \centering
    \small
\begin{tcolorbox}[
  colback=blue!5,
  colframe=blue!60!black,
  title=Prompt of the External LLM Judge (Part Two)
]
These examples illustrate the pattern where the model performs reflection on intermediate parts of the reasoning before a final answer is derived. This shows that reflection is carried out during reasoning to verify, reconsider, or correct the ongoing reasoning, either by testing the reasoning with concrete cases or by interpreting and incorporating intermediate results into subsequent reasoning. These examples are illustrative and do not define all possible cases.

\paragraph{Category 3: No reflection.}
The model performs reasoning and derives a final answer without performing any reflection, either during the reasoning or after the final answer is obtained. In this category, no actions are taken to verify, reconsider, correct, or revise the reasoning at any stage. Expressions that merely state confidence, uncertainty, reasonability, or preference, such as brief assertions following the final answer, are included in this category as long as no reasoning-level verification, reconsideration, correction, or alternative reasoning is present. Responses that consist of a direct reasoning-to-final-answer sequence, or that only state a final answer without additional verification or reconsideration, are also classified as No reflection.

Here is one illustrative example:

Let me compute that:

First, compute $324^2$:

\ldots

So $R = 540$. Therefore, the maximum value of the real part is $540$.

So the answer is $540$. I think I can confidently say that the largest possible real part is $540$.

\textbf{Final Answer}

The largest possible real part is $\boxed{540}$.

This example illustrates the pattern where the model proceeds directly from reasoning to a final answer without performing any reflection. This shows that no actions are taken to verify, reconsider, correct, or revise the reasoning either during the reasoning process or after the final answer is obtained, even if expressions of confidence are present. This example is illustrative and does not define all possible cases.

\paragraph{Additional classification rules.}
\begin{itemize}
    \item If reflection occurs both during reasoning and after the final answer, return both Category 1 and Category 2.
    \item If the response contains no reasoning-level verification, reconsideration, correction, or alternative reasoning, return Category 3.
    \item If the response is empty, incomplete, or only gives a final answer without verification or reconsideration, return Category 3.
\end{itemize}

\paragraph{Response to evaluate.}
\begin{verbatim}
{{MODEL_RESPONSE}}
\end{verbatim}

\paragraph{Output format.}
Return the result as JSON with the following field:
\begin{itemize}
    \item \texttt{categories}: a list of one or more category names.
\end{itemize}

Do not add explanations or extra fields.
\end{tcolorbox}
\caption{An example prompt used for external LLM judge (Part Two).}
    \label{fig:judge_prompt_2}
\end{figure*}


\end{document}